# A Deep Learning Inference Scheme Based on Pipelined Matrix Multiplication Acceleration Design and Non-uniform Quantization

**Yuyang Zhang, Dik Hin Leung\*, Min Guo, Yijia Xiao,
Haoyue Liu, Yunfei Li, Jiyuan Zhang, Guan Wang, Zhen Chen**

Tsinghua University, Beijing 100084, China
*Corresponding author:* \* 1dx18@mails.tsinghua.edu.cn

**Abstract:** Matrix multiplication is the bedrock in Deep Learning inference application. When it comes to hardware acceleration on edge computing devices, matrix multiplication often takes up a great majority of the time. To achieve better performance in edge computing, we introduce a low-power Multi-layer Perceptron (MLP) accelerator based on a pipelined matrix multiplication scheme and a nonuniform quantization methodology. The implementation is running on Field-programmable Gate Array (FPGA) devices and tested its performance on handwritten digit classification and Q-learning tasks. Results show that our method can achieve better performance with fewer power consumption.

## 1 Introduction

Neural networks (NN) and deep learning techniques have been widely applied in the areas of image classification, computer vision, natural language processing, etc. There are considerable efforts have been made in accelerating the inference process of deep learning tasks, which involves calculating the output of neural network (NN) models under given inputs. The three most commonly used neural network types are: (sparse) Convolutional Neural Networks (CNN), and Recurrent Neural Networks (RNN), and Multi-Layer Perceptron (MLP). Previous works on NN accelerators mainly focused on CNN and RNN models, which are widely used in CV and NLP tasks.

Comparing with other machine learning techniques (e.g. Linear Regression and SVM), MLP is the fundamental structure, but proved to be useful in fitting all kinds of functions, both theoretically and empirically. In practice, MLP is widely used in multivariate classification and regression tasks. Another application scenario is the reinforcement learning (RL) algorithms, which use MLPs as function approximators in value-based reinforcement learning (RL) algorithms such as Q-learning.

In practice, when we deploy the NN model in real-time downstream tasks like anomaly detection, we expect the model to generate response to input data as fast as possible. A popular solution is to carry out inference process in an edge device that is closer to source of the data (e.g. a camera) than in cloud, which is called inference at the edge. There are examples where edge inference is used in control tasks like robot control and low power budget drone control. There are also other application scenarios for edge inference on MLP based models, such as network traffic classification and monitoring, sensor on-edge classification on IoT devices, etc.

When deploying deep learning models for inference at the edge, such as, evaluating RL control algorithms on drones. MLP models are used in edge inference where power are rather limited, while the inference speed are restricted for lower energy consumption by limited power input. For example, although satisfactory inference speed can be reached with GPU, the whole solution is too expensive and extremely power-consuming. In this case, FPGA is the choice in balancing the inference speed and the power consumption.

From above, we can see the importance of reducing power usage of MLP accelerators in edge inference. In this paper, we propose a novel framework of scalable low-power circuit-based MLP accelerator and its FPGA implementations.

The structure of the paper is organized as follows:
- In Section 2, we briefly recap the three main paradigms of MLP acceleration.
- In Section 3, we introduce our methods, using a pipelined matrix multiplication framework with input buffer to decompose data loading and computation.
- In Section 4, we provide the result of our experiment of acceleration on handwritten digit classification.
- In Section 5, we conclude the whole paper.

## 2 Related Works

As one of the most basic NN models in deep learning, the multi-layer perceptron is widely used in CNN, RNN, and Attention mechanism models. Therefore, by optimizing the multi-layer perceptron model, we can speed up the inference process of NN models and reduce the power consumption. From the perspective of linear algebra, the inference process for an input on multi-layer perceptron model is an alternating stack of matrix multiplication and activation functions. Therefore, there are many optimizations for the calculation of matrix multiplication, which can be categorized into three main paradigms.

## 2.1 Optimization based on input characteristics

Not all neural networks require GEMM (General Matrix Multiplication). In some cases, researchers can leverage the data patterns of matrix, which can be applied into the optimization of special matrices, such as the optimization of boolean matrices, the optimization of matrices with special laws (block matrices), and the optimization of sparse matrices from the level of circuit design. For example, in order to reduce the computational cost of block matrix multiplication, we can divide the original matrix into submatrices, and do multiplication only for non-zero ones. Such technique improves the efficiency of calculation, and reduces power consumption.

## 2.2 Optimization based on mixed precision

Since the accuracy of a single value does not affect the overall calculation result in deep learning, researchers have developed methods of mixed precision, speeding up calculations by using high-precision representations of numbers only when necessary and reduced precision for the others. For example, Intel APEX supports fp16 calculations, which can reduce the bit length of floating-point numbers to half of the original.

## 2.3 Optimization based on quantitative methods

Quantification method[5] works by approximating the result of the neural network. Although accuracy may be lost in the process of quantification, the approximated result can still reach the needs when certain computing resources are scarce. For example, in some embedded systems, some floating-point numbers are converted into integers during inference.

## 3 Proposed pipeline matrix-multiply acceleration framework on FPGA

For accelerating the matrix multiplication in FPGA, we design a pipelined matrix multiplication framework, where the input buffers are decomposed for data loading and computation, following similar intuition idea with processing in memory (PIM). Moreover, a nonuniform quantization algorithm is extended and implemented to this setting for hardware-friendly computation.

### 3.1 Pipeline matrix-multiply framework

A new Dataflow framework is designed as shown below:

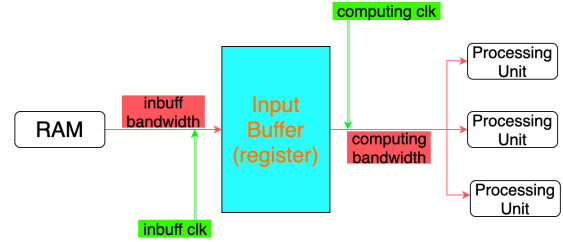

**Figure 1** Input Buffer loads data from RAM at the speed of $bandwidth_{inbuf}$, controlled by clock $clk_{inbuff}$.

When preprocessing data for weight ($m \times n$)-data ($n \times 1$) multiplication $w \cdot d$, the weight matrix is first decomposed into m rows $w_1, w_2, ..., w_m$ with size $n \times 1$. Then every weight row ($1 \times n$) is concatenated with the data ($1 \times n$) to form m reorganized data rows with size $2n \times 1$, which is inspired by the data organizing method by Sudrajat[5]. All of them are loaded into the input buffer.

After preprocessing, every reorganized data row is fed into first-level Processing Units (PUs). For the next several clock cycles, quantized float multiplication and addition is executed following the process shown below. Meanwhile, descendent data flows in and follows the same procedure mentioned above, but always a clock cycle behind previous data, forming the pipelined matrix-multiplication framework. Finally, at the end of every clock cycle t, we get the dot product of $w_i \cdot d_t$ as outputs of the first-level PUs. These outputs are concatenated horizontally, forming the result of $w \cdot d_t$.

The framework presented above not only enabled pipelined computation of weight-data multiplication, but also decompose the data-loading process and the data-computing process. Note that all of the data flowed into the input buffer is controlled by the clock-cycle of input buffer writing $clk_{inbuff}$ while the data bandwidth is affected by both $clk_{inbuff}$ and the bandwidth of RAM-buffer communication. However, data computing is controlled by an asynchronized clock $clk_{compute}$. The decomposition is feasible thanks to the asynchronization of data loading and computing as long as data loading is faster than data computing. We argue that the data loading can be significantly faster because of the large bandwidth although the data loading clock-cycle is necessarily larger than the computing clock-cycle. For example, data loading may require a clock-cycle larger than 300ns, but the data loaded within 300ns may require 500ns to be computed.

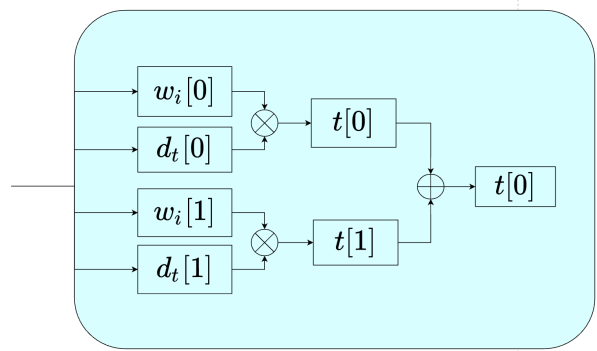

**Figure 2** $w_i$ is the ith row of the weight matrix, $d_t$ is the data at time t, array t is the temporary result.

### 3.2 Extended Power-of-Two (PoT) quantization

Quantization is extensively used to decrease computation, including uniform and nonuniform methodology.

#### A. Uniform quantization

Uniform quantization corresponds to uniform interval when mapping continuous floating-point values into low bit figures. The generally used binary or ternary methodology maps continuous values into {0, 1} or {-1, 0, 1}. Under these mapping, multiplications are reduced to simple logic manipulations like AND and OR. Although these mappings seem to bear low precision at first glance, recent works have proven acceptable accuracy loss.

There are also other low-bit quantization methodologies, but they all follow the same intuition under the setting of uniform quantization.

#### B. Nonuniform quantization

A representative nonuniform quantization is the Power-of-Two (PoT) quantization. A typical b-bit quantization in $[-\alpha, \alpha]$ can be shown as

$$Q(b, \alpha) = \alpha \times \left\{0, \pm \frac{1}{2^{2^{b-1}-1}}, \pm \frac{1}{2^{2^{b-1}-2}}, \dots, \pm \frac{1}{2}, \pm 1\right\} \quad (3.1)$$

By PoT quantization, multiplication is reduced into the following form:

$$2^m \times q = \begin{cases} q \ll |m|, m < 0 \\ q \gg m, m > 0 \end{cases} \quad (3.2)$$

The PoT method, however, suffer from low-accuracy at the tail ends of the quantization interval. This is due to the sparsity of quantization level at these two ends.

A novel quantization method called SP2 was recently developed by Chang[1] to relieve this difficulty. The b-bit quantization is represented as follow:

$$Q(b, \alpha) = \pm \alpha \times \{q_1 + q_2\} \quad (3.3)$$

$$q_1 \in \left\{0, \pm \frac{1}{2^{2^{b_1}-1}}, \pm \frac{1}{2^{2^{b_1}-2}}, \dots, \frac{1}{2}\right\}$$

$$q_2 \in \left\{0, \pm \frac{1}{2^{2^{b_2}-1}}, \pm \frac{1}{2^{2^{b_2}-2}}, \dots, \frac{1}{2}\right\}$$

where $b_1 + b_2 = b - 1$. The quantization level generated this way can bear more linear identity near the two tail ends where the weights may distribute, which greatly increase accuracy.

In our setting, we further extend this algorithm into SPx quantization as follow so as to gain more flexible attributes at these two ends:

$$Q(b, \alpha) = \pm \alpha \times \sum_i q_i \quad (3.4)$$

$$q_i \in \left\{0, \pm \frac{1}{2^{2^{b_i}-1}}, \pm \frac{1}{2^{2^{b_i}-2}}, \dots, \frac{1}{2}\right\}$$

$$b = \sum_i b_i$$

Intuitively speaking, if we extend 2 to x, we can get more choices at the two tail ends of the quantization interval. Although more computation is involved, more flexibility is achieved through the trade off between computation and better performance at the two tail ends.

## 4 Experiment

### 4.1 Neural Network Framework

The following defines an MLP model formally.

Assume the MLP consists of $N$ layers, including input layer and output layer, and $N_i (1 \leq i \leq n)$ is the number of neurons in the $i$-th layer. Let $F_i(x; \theta)$ be the output vector of the $i$-th layer in the MLP with parameters $\theta$. Define:

$$F_i(x; \theta) = \begin{cases} \sigma^{(1)}(x + b^{(1)}) & i = 1 \\ \sigma^{(i)}(W^{(i)} F_{i-1}(x; \theta) + b^{(i)}) & 2 \leq i \leq N \end{cases} \quad (4.1)$$

where $\sigma^{(i)}$ is the activation function of the $i$-th layer, $W^{(i)} \in R^{N_i \times N_{i-1}}$ is the weight matrix between layer $i-1$ and layer $i$, and $b^{(i)} \in R^{N_i}$ is the bias vector of layer $i$. The final output of the network with input $x$ is $F_N(x; \theta)$.

In this experiment setting, we use an MLP model where $N = 3$. The input layer consists of 784 neurons, the hidden layer 128 neurons, and the output layer 10. The activation function of the hidden layer and the output layer is sigmoid function (i.e. $\sigma(x) = \frac{1}{1+e^{-x}}$). Thus, the final output $F(x; \theta)$ with input $x$ of our network can be described as:

$$F(x; \theta) = \sigma(W^{(3)} \sigma(W^{(2)} x + b^{(2)}) + b^{(3)}) \quad (4.2)$$

To obtain the final classification result $y \in \{0,1,2,\dots,9\}$

from the output, we set

$$y = arg \max_{y \in \{0,1,\ldots,9\}} F^{(y)}(x;\theta) \quad (4.3)$$

where $F^{(y)}(x;\theta)$ is the $y$-th component of the vector $F(x;\theta)$.

The parameters $\theta$ are pretrained on a CPU or GPU via stochastic gradient descent (SGD) algorithm. Denote the training set with $D = \{(x_i, y_i)\}_{i=1}^{M}$, where $x_i$ and $y_i$ denote the $i$-th data point and its label, and $M$ denotes the size of the dataset. We use mean square error loss function $L(x;\theta)$, which is defined as:

$$L(x;\theta) = \frac{1}{M} \sum_{i=1}^{M} |F(x_i;\theta) - Y_i|_2^2 \quad (4.4)$$

where $Y_i$ is the one-hot vector consists of 10 components, with the $y_i$-th component set to 1.

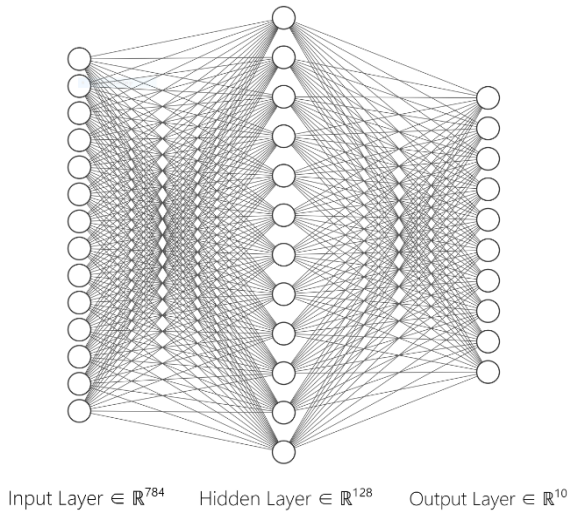

**Figure 3** A simplified diagram of the MLP model

In each update step, we sample a mini-batch $S \subseteq D$, whose size is equal to batch size $B$, a hyperparameter in training process. We estimate the full loss function with Eq.(4.5).

$$\hat{L}(x;\theta) = \frac{1}{B} \sum_{(x_i,y_i) \in S} |F(x_i;\theta) - Y_i|_2^2 \quad (4.5)$$

Denoting the learning rate with $\eta$, we update the parameters $\theta$ with the following rules Eq.(4.6):

$$\theta' \leftarrow \theta - \eta \frac{\partial \hat{L}(x;\theta)}{\partial \theta} \quad (4.6)$$

In out experiment, we set $B = 64, \eta = 0.5$.

### 4.2 Reinforcement Learning Experiment

Furthermore, we conduct a series of experiments to test the performance of inference in NN in the Acrobot-v1 environment of OpenAI Gym, where we use an MLP model as a Q function approximator in Q-learning algorithm.

### 4.3 Dataset used in Experiment

The handwritten digit recognition task is selected as the application scenario based on MNIST dataset, which contains a total of 60,000 ($28 \times 28$) pixel images. The grayscale values of the 784 pixels in total, function as the input of the MLP neural network.

### 4.4 Experiment Result Summary

In general, we measure the power consumption and execution time for different edge devices. The whole power consumption measure system consists of a CPU platform which can plugged-in a GPU or FPGA platform.

The procedure of our experiment can be seen as follows:

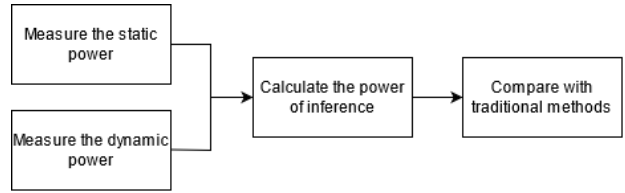

**Figure 4** method of power measure and comparison

Detailed experimental methods are shown as follows:

A. CPU

We firstly measure the standby power consumption and record the time when neural network starts inference, then we measure the dynamic power consumption when inference is in process, recording the time when it is finished. The time and power consumption of inference in CPU can be calculated in this way.

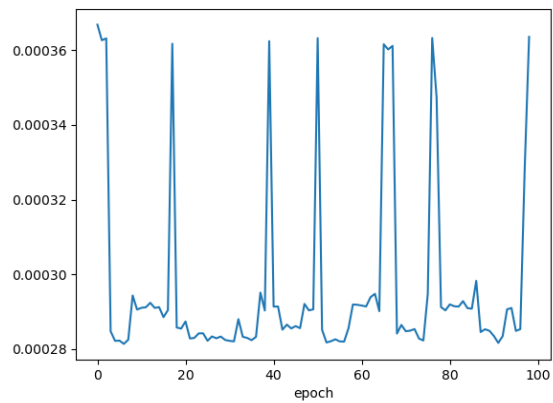

**Figure 5.** The measured inference time per sample (x-axis labels the epochs, y-axis labels time(sec) per sample)

The detailed measure of inference time is shown in Figure 5. We measure the whole inference time with a Batch size of samples as input, and then we can calculate the inference time of an individual sample.

B. GPU

Similarly, we measure the inference time with two configuration, i.e. standby power consumption and dynamic powering consumption. What is the different here is that the final GPU power consumption of our model will be the working consumption minas the standby power consumption of CPU.

C. FPGA

Verilog Implement of the circuit design of MLP with FPGA board(Intel APEX), we then measure the inference time per sample and the power consumption of the whole inference process.

The experiment results are as shown in Table 1 as follows:

**Table I** Comparison of CPU, GPU and FPGA inference time per sample and total power consumption on handwritten digits recognition task

|      | Time per sample (s) | Power consumption (W) |
| --- | --- | --- |
| **CPU** | $2.6 \times 10^{-3}$ | 47.2 |
| **GPU** | $3 \times 10^{-4}$ | 115.2 |
| **FPGA** | $1.6 \times 10^{-6}$ | 10 |

From the results shown in Table I, FPGA based MLP design outperforms common existing general computing devices, like CPU and GPU. However, due to limited experiment conditions, a more accurate approximation of the power consumption is needed, which is left for the future research.

## 5 Conclusion

This paper proposed a MLP accelerator based on FPGA, which utilizes quantization and a novel matrix multiplication design. The matrix multiplication design therein decomposes data-loading process (from RAM) apart from data-computing process (within registers) thanks to our input buffer, which enabls faster data computing without the limit of the slow data-loading process. Moreover, we split matrix multiplication into scalar multiplication and addition, which can be further fit into a pipeline design, largely accelerating the data computing. Meanwhile, we also implement floating-point quantization on FPGA by using the extended SP-x quantization methodology, which simplify scalar calculation while achieving good linear characteristics in almost the whole quantization interval. Lastly, we tested our design both on the extensively used MNIST dataset and in Q-learning algorithm, receiving better performance with fewer power consumption.